\documentclass[a4paper,twoside]{article}

\usepackage{epsfig}
\usepackage{subcaption}
\usepackage{calc}
\usepackage{amssymb}
\usepackage{natbib}
\usepackage{amstext}
\usepackage{amsmath}
\usepackage{amsthm}
\usepackage{multicol}
\usepackage{url}
\usepackage{pslatex}
\usepackage{apalike}
\usepackage{algorithm2e}
\usepackage[bottom]{footmisc}
\usepackage{SCITEPRESS}     

\usepackage[colorlinks=true, linkcolor=blue, citecolor=blue]{hyperref}
\usepackage[capitalise,noabbrev,nameinlink]{cleveref}
\usepackage{tikz}
\newtheorem{definition}{Definition}

\usetikzlibrary{automata, positioning}
\usetikzlibrary{shapes.geometric, arrows, positioning}

\tikzstyle{block} = [rectangle, draw, text centered, rounded corners, minimum height=2em]
\tikzstyle{line} = [draw, -stealth, thick]
\tikzstyle{cloud} = [ellipse, draw, text centered, minimum height=2em, thick]
\tikzstyle{dashedcloud} = [ellipse, draw, dashed, text centered, minimum height=2em, thick]
\usetikzlibrary{positioning, arrows.meta, shapes.geometric, decorations.pathreplacing}

\tikzstyle{startstop} = [rectangle, rounded corners, minimum width=1.5cm, minimum height=0.5cm,text centered, draw=black, fill=red!30]
\tikzstyle{io} = [trapezium, trapezium left angle=70, trapezium right angle=110, minimum width=1cm, minimum height=0.5cm, text centered, draw=black, fill=blue!30]

\tikzstyle{process} = [rectangle, minimum width=3cm, minimum height=0.5cm, text centered, draw=black, fill=orange!30]
\tikzstyle{decision} = [diamond, minimum width=0.5cm, minimum height=0.1cm, text centered, draw=black, fill=green!30]
\tikzstyle{process2} = [rectangle, minimum width=1cm, minimum height=0.5cm, text centered, draw=black, fill=orange!30]
\tikzstyle{arrow} = [thick,->,>=stealth]
\usetikzlibrary{shapes,shadows,arrows,positioning,angles,quotes}
\tikzset{My Arrow Style/.style={single arrow, fill=black!15, anchor=base, align=center,text width=2.3cm}}
\tikzstyle{arrow} = [thick,->,>=stealth]
\usetikzlibrary{calc,trees,positioning,arrows,fit,shapes,calc}

\begin{document}

\title{Co-Activation Graph Analysis of Safety-Verified and Explainable Deep Reinforcement Learning Policies}


\author{\authorname{Dennis Gross, and Helge Spieker}
\affiliation{Simula Research Laboratory, Norway}
\email{dennis@simula.no}
}
\abstract{\emph{Deep reinforcement learning (RL)} policies can demonstrate unsafe behaviors and are challenging to interpret.
To address these challenges, we combine RL policy model checking—a technique for determining whether RL policies exhibit unsafe behaviors—with co-activation graph analysis—a method that maps neural network inner workings by analyzing neuron activation patterns—to gain insight into the safe RL policy's sequential decision-making.
This combination lets us interpret the RL policy's inner workings for safe decision-making.
We demonstrate its applicability in various experiments.}
\onecolumn \maketitle \normalsize \setcounter{footnote}{0} \vfill

\section{\MakeUppercase{Introduction}}
\emph{Deep Reinforcement Learning (RL)} has improved various industries~\citep{DBLP:journals/ress/LiuYY24,DBLP:journals/ijpr/JiLXYLZ24,DBLP:journals/artmed/WangLYWXCW24}, enabling the creation of agents that can outperform humans in sequential decision-making tasks~\citep{mnih2015human}.

In general, an RL agent aims to learn a near-optimal policy to achieve a fixed objective by taking actions and receiving feedback through rewards and state observations from the environment~\citep{sutton2018reinforcement}.
Each state is described in terms of features, which can be considered characteristics of the current environment state~\citep{strehl2007efficient}.
We call a policy a \emph{memoryless policy} if it only decides based on the current state~\citep{sutton2018reinforcement}.

A \emph{neural network (NN)} commonly represents the policy that, given the observation of the environment state as input, yields values that indicate which action to choose~\citep{mnih2013playing}.
These values are called \emph{Q-values}~\citep{watkins1992q}, representing the expected cumulative reward an agent policy expects to obtain by taking a specific action in a particular state.

Unfortunately, trained policies can exhibit \emph{unsafe behavior}~\citep{DBLP:conf/setta/GrossJJP22} like collisions~\citep{DBLP:journals/access/BanL24}, as rewards often do not fully capture complex safety requirements~\citep{DBLP:journals/aamas/VamplewSKRRRHHM22}, and are \emph{hard to interpret} because the complexity of NNs hides crucial details affecting decision-making~\citep{DBLP:journals/ml/Bekkemoen24}.

To resolve the issues mentioned above, formal verification methods like \emph{model checking}~\citep{baier2008principles} have been proposed to reason about the safety of RL policies~\citep{yuwangPCTL,DBLP:conf/formats/HasanbeigKA20,DBLP:conf/atva/BrazdilCCFKKPU14,DBLP:conf/tacas/HahnPSSTW19} and \emph{explainable RL methods} to interpret trained RL policies~\citep{DBLP:journals/csur/MilaniTVF24}.

Model checking is not limited by the properties that rewards can express.
Instead, it supports a broader range of properties that can be expressed by \emph{probabilistic computation tree logic (PCTL)}~\citep{DBLP:journals/fac/HanssonJ94}.
PCTL formalizes reasoning about probabilistic systems, such as \emph{Markov decision processes (MDPs)}. It enables the specification of (safety) properties that relate to the probability of events occurring over discrete time steps, as applicable in our RL setting.

Explainable RL involves methods that make RL policies interpretable, such as clarifying how the policy makes decisions~\citep{DBLP:conf/aiide/SieusahaiG21}.
Local explanations clarify decision-making for specific states, while global explanations offer a holistic view of the policy and its actions~\citep{DBLP:journals/csur/MilaniTVF24}.

Some research combines safety with explainability by creating simpler surrogate models of policies~\citep{DBLP:conf/nips/0001LDL23}, pruning neural network interconnections and re-verifying the pruned network to identify which connections influence safety properties~\citep{SafetyPruningESANN2024}.
Other approaches use external systems to explain failures and propose alternative actions, enhancing the safety of trained RL policies~\citep{gross2024enhancing}.

Unfortunately, there remains a \emph{gap between local and global explanations}, as, to the best of our knowledge, no current methodology offers nuanced \emph{safety explanations} for RL policies within specific regions of the environment.

\emph{Co-activation graph analysis}~\citep{DBLP:journals/fgcs/HortaTLM21,DBLP:conf/kcap/SelaniT21,DBLP:conf/nesy/HortaSSM23,DBLP:conf/aiia/HortaM21} can be such methodology.
While co-activation graph analysis was successfully applied in classification tasks, no work has applied it to RL nor in the context of safety.

In general, co-activation graph analysis explores how NN classifiers learn by extracting their acquired knowledge~\citep{DBLP:journals/fgcs/HortaTLM21}.
The method creates a graph in which the nodes represent neurons, and the weighted connections show the statistical correlations between their activations.
Correlations are derived by applying the trained NN classifier to a labeled dataset (labels come from external knowledge) and measuring the relationships between neuron activations.

However, the main challenge in RL is identifying and integrating the missing external knowledge into the co-activation graph analysis to extract valuable information from the trained NN policies.

In this work, we tackle the problem of generating external knowledge via safety verification and explainable RL methods to allow co-activation graph analysis in the context of RL safety. 
This approach creates a new category of explainable RL methods, which we call \emph{semi-global safety explanations}.
We achieve this through the following steps.

First, we create the unlabeled dataset containing the states of the environment that are reachable by the trained RL policy and for which a user-specific safety property holds.
In more detail, given a model-based RL environment, a user-specified safety property, and a trained RL policy, the formal model of the interactions between the RL environment and trained RL policy is built and verified in the following way.
We query for an action for every state reachable via the trained policy relevant to the given safety property.
Only states reachable via that action are expanded in the underlying environment.
The resulting formal model is fully deterministic, with no open action choices.
It is passed to the model checker Storm for verification, yielding the \emph{exact} safety property and all its relevant states~\cite{DBLP:conf/setta/GrossJJP22}.

Second, we label the whole state dataset with the safety property as the label and compare it with other labeled state datasets (such as another safety property labeled dataset), or we label each state individually in the dataset via an explainable metric (for instance, if the state is critical or not critical for the trained policy~\citep{DBLP:journals/csur/MilaniTVF24,vouros_explainable_2023}) or another user-specified metric.

Finally, we investigate the neuron activations of the trained RL policy for the labeled datasets via co-activation graph analysis methods~\citep{DBLP:journals/fgcs/HortaTLM21,DBLP:conf/kcap/SelaniT21,DBLP:conf/nesy/HortaSSM23,DBLP:conf/aiia/HortaM21} to gain insights into the trained NN policy inner-workings by analyzing the neuron co-activations per labeled dataset and compare.

Our experiments show that RL co-activation graph analysis is a valuable tool for interpreting NNs in RL policies, especially for safety applications.
It offers insights into neuron importance and feature rankings, and it identifies densely connected neuron clusters, or functional modules, within the network.
This reveals how different parts of the neural network contribute to safe decision-making, enhancing our understanding of the model's behavior in critical areas and fueling human curiosity in the pursuit of explainable AI~\citep{DBLP:journals/fcomp/HoffmanMKL23,DBLP:journals/corr/abs-1812-04608,miao2018humanized}.

Therefore, our \textbf{main contribution} is a framework that allows us to apply co-activation graph analysis specifically for RL safety interpretations.

\section{Related Work}
In this section, we review work related to our approach.
First, we position our method within the broader field of explainable techniques for NNs.
Next, we examine research focused on the formal verification of RL policies.
Finally, we discuss studies integrating explainability with formal verification of RL policies, highlighting where our approach contributes within this combined framework.

\subsection{Explainable NN Methods}
Drawing inspiration from neuroscience, which uses network analysis and graphs to understand the brain, \citet{DBLP:journals/fgcs/HortaTLM21} explore how NNs learn by extracting the knowledge they have acquired.
They developed a co-activation graph analysis in the context of classification tasks.
The authors suggest that this co-activation graph reflects the NN's knowledge gained during training and can help uncover how the NN functions internally.
In this graph, the nodes represent neurons, and the weighted connections show the statistical correlations between their activations.
These correlations are derived by applying the trained NN classifier to a labeled dataset (labels come from external knowledge) and measuring the relationships between neuron activations.
This method enables identifying, for instance, the most important neurons for classifying a specific class or which features are essential for a specific class type.

Building upon the foundational work by \citet{DBLP:journals/fgcs/HortaTLM21} on co-activation graphs in classification tasks, \citep{DBLP:conf/kcap/SelaniT21} uses co-activation graph analysis to investigate it in the context of autoencoders used in anomaly detection.

Further work exists combining explainable artificial intelligence with co-activation graph analysis~\citep{DBLP:conf/nesy/HortaSSM23,DBLP:conf/aiia/HortaM21}.
Their co-activation graph analysis focuses on providing textual explanations for convolutional NNs in image classification tasks by connecting neural representations from trained NNs with external knowledge, using the co-activation graph to predict semantic attributes of unseen data, and then generating factual and counterfactual textual explanations for classification mistakes.

We extend the branch of co-activation graph research by setting co-activation graph analysis in the context of RL.
The challenge is identifying and integrating the missing external knowledge into the co-activation graph analysis to extract valuable information from the trained NN policies.

In RL, local explanations focus on why an RL policy selects a particular action at a specific state~\citep{DBLP:journals/csur/MilaniTVF24}.
Our approach extends these local explanations into larger state sets by creating the co-activation graph dataset using RL policy model checking, combining explainability with safety.
Compared to global explanations that aggregate the overall policy behavior, co-activation allows us to get a more fine-grained analysis of the original trained RL policy.
Leading to so-called \emph{semi-global explanations}.

\subsection{Formal Verification of RL Policies}
Various studies use model checking to verify that RL policies do not exhibit unsafe behavior~\citep{DBLP:conf/sigcomm/EliyahuKKS21,DBLP:conf/sigcomm/KazakBKS19,pmlr-v161-corsi21a,DBLP:journals/corr/DragerFK0U15,DBLP:conf/pldi/ZhuXMJ19,DBLP:conf/seke/JinWZ22,DBLP:conf/setta/GrossJJP22}.
We build on top of the work of \citet{DBLP:conf/setta/GrossJJP22} and augment their tool to support co-activation  RL policy graph analysis~\citep{DBLP:conf/concur/CassezDFLL05,DBLP:conf/tacas/DavidJLMT15}.

\subsection{Formal Verification and Explainability}
In the context of MDPs, there exists work~\citep{DBLP:conf/exact/ElizaldeSRd07,DBLP:conf/micai/ElizaldeSNR09} that analyzes the feature importance (a type of explainability) of MDPs manually and automatically.
However, we focus on the inner workings of \emph{trained RL policies} for states that satisfy user-specified safety properties.

In the context of \emph{classification tasks}, work exists that extends the PCTL language by itself to support more trustworthiness of explanations~\citep{DBLP:conf/lori/TerminePD21}.
However, we focus on sequential decision-making of RL policies.

In the context of explainable and verified RL, existing work iteratively prunes trained NN policies to interpret the feature importance for safety at a global level~\citep{SafetyPruningESANN2024}.
We support various graph algorithms applied to the inner workings of NN policies, including a way to measure the feature importance.
Additionally, some work leverages large language models to identify safety-critical states and apply counterfactual reasoning to explain why the RL policy violated a safety property while proposing alternative actions~\citep{gross2024enhancing}.
However, this approach provides only local~explanations via an external large language model.

\section{Background}
First, we introduce probabilistic model checking.
Second, we present the basics for explainable RL.
Finally, we give an introduction to co-activation graph~analysis.

\subsection{Probabilistic Model Checking}
A \textit{probability distribution} over a set $X$ is a function $\mu \colon X \rightarrow [0,1]$ with $\sum_{x \in X} \mu(x) = 1$. The set of all distributions on $X$ is $Distr(X)$.

\begin{definition}[MDP]\label{def:mdp}
A \emph{MDP} is a tuple $M = (S,s_0,Act,Tr, rew,$\\$AP,L)$
where $S$ is a finite, nonempty set of states; $s_0 \in S$ is an initial state; $Act$ is a finite set of actions; $Tr\colon S \times Act \rightarrow Distr(S)$ is a partial probability transition function;
$rew \colon S \times Act \rightarrow \mathbb{R}$ is a reward~function;
$AP$ is a set of atomic propositions;
$L \colon  S \rightarrow 2^{AP}$ is a labeling function.
\end{definition}
We employ a factored state representation where each state $s$ is a vector of features $(f_1, f_2, ...,f_d)$ where each feature $f_i\in \mathbb{Z}$ for $1 \leq i \leq d$ (state dimension).
\begin{definition}
    A \emph{memoryless deterministic policy $\pi$} for an MDP $M$ is a function $\pi \colon S \rightarrow Act$ that maps a state $s \in S$ to action $a \in Act$.
\end{definition}
Applying a policy $\pi$ to an MDP $M$ yields an \emph{induced DTMC} $D$ where all non-determinism is resolved.

Storm~\citep{DBLP:journals/sttt/HenselJKQV22} is a model checker. 
It enables the verification of properties in induced DTMCs, with reachability properties being among the most fundamental.
These properties assess the probability of a system reaching a particular state.
For example, one might ask, ``Is the probability of the system reaching an unsafe state less than 0.1?''
A property can be either \emph{satisfied} or \emph{violated}.

\begin{figure}
    \centering
    \scalebox{0.8}{
    \begin{tikzpicture}[node distance=0.5cm, auto]
        \node [cloud] (system) {system};
        \node [cloud, below=0.5cm of system] (systemdesc) {system description};
        \node [cloud, below=0.5cm of systemdesc] (model) {model};
        \node [cloud, right=0.15cm of systemdesc] (requirements) {requirements};
        \node [cloud, below=0.5cm of requirements] (properties) {properties};
        \node [block, below=0.5cm of model] (modelchecking) {model checking};
        \node [cloud, below left=0.15cm of modelchecking] (satisfied) {satisfied};
        \node [cloud, below right=0.15cm of modelchecking] (violated) {violated};
    
        \path [line] (system) -- node[right] {modeling} (systemdesc);
        \path [line] (systemdesc) -- node[right] {translates to} (model);
        \path [line] (requirements) -- node[right] {formalizing} (properties);
        \path [line] (model) -- (modelchecking);
        \path [line] (properties) -- (modelchecking);
        \path [line] (modelchecking) -- (satisfied);
        \path [line] (modelchecking) -- (violated);
    
    \end{tikzpicture}
    }
    \caption{Model checking workflow~\citep{DBLP:journals/sttt/HenselJKQV22}. First, the system needs to be formally modeled, for instance, via PRISM. Then, the requirements are formalized, for instance, via PCTL. Eventually, both are inputted into the model checker, like Storm, which verifies the property.}
    \label{fig:model_checking}
\end{figure}
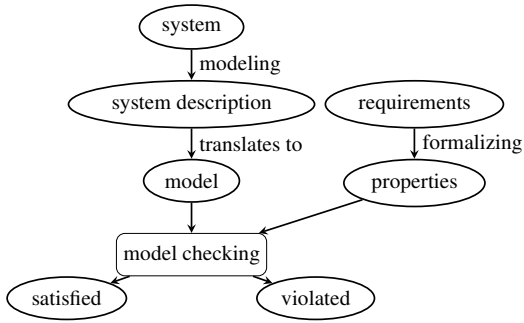

The \emph{general workflow} for model checking with Storm is as follows (see Figure~\ref{fig:model_checking}):
First, the system is modeled using a language such as PRISM~\citep{prism_manual}.
Next, a property is formalized based on the system's requirements.
Using these inputs, the model checker Storm verifies whether the formalized property is satisfied or violated within the model.

In probabilistic model checking, there is no universal ``one-size-fits-all'' solution~\citep{DBLP:journals/sttt/HenselJKQV22}.
The most suitable tools and techniques depend significantly on the specific input model and properties being analyzed.
During model checking, Storm can proceed ``on the fly'', exploring only the parts of the DTMC most relevant to the verification.

\subsection{Explainable Reinforcement Learning}
The standard learning goal for RL is to learn a policy $\pi$ in an MDP such that $\pi$ maximizes the accumulated discounted reward~\citep{DBLP:journals/ml/Bekkemoen24}, that is, $\mathbb{E}[\sum^{N}_{t=0}\gamma^t R_t]$, where $\gamma$ with $0 \leq \gamma \leq 1$ is the discount factor, $R_t$ is the reward at time $t$, and $N$ is the total number of steps.

To approximate the optimal policy $\pi^*$ concerning the objective, RL algorithms employ NN, which contains multiple layers of neurons, as function approximators~\citep{mnih2013playing}.

Explainability methods are used to understand trained RL policies~\citep{DBLP:journals/csur/MilaniTVF24}.
Global explainable RL methods, for instance, build understandable surrogate policies of the original policy that are better understandable, but perform less well~\citep{DBLP:conf/aiide/SieusahaiG21}.
Local explanation methods explain the decision-making of a policy in a given environment state.

\paragraph{Critical state} A common approach to explanations in RL is to highlight the most critical states in a trajectory~\citep{DBLP:journals/csur/MilaniTVF24,vouros_explainable_2023}, i.e., those states where the choice of action has a large impact on the accumulated rewards of the episode. 
A measure that has been used to locate critical states from the output of policies is the \textit{state importance}~\citep{DBLP:conf/atal/TorreyT13,huang_establishing_2018}.
The state importance judges a state's relevance by the policy network's outputs, i.e., the scores the policy assigns to each action. For instance, the distance between the highest and lowest scores above a threshold can be considered critical.

\begin{figure}[]
\centering
\scalebox{1}{
    \begin{tikzpicture}[]
     {};
    \node (agent1) [process] {RL Agent};
    \node (env) [process, below of=agent1,yshift=-0.25cm,xshift=2cm] {Environment};
    
    \draw [arrow] (agent1) -| node[anchor=west] {Action} (env);
    \draw [arrow] (env) -| node[anchor=east] {New State, Reward} (agent1);
    \end{tikzpicture}
}
\caption{This diagram represents an RL system in which an agent interacts with an environment. The agent receives a state and a reward from the environment based on its previous action. The agent then uses this information to select the next action, which it sends to the environment.}
\label{fig:rl}
\end{figure}
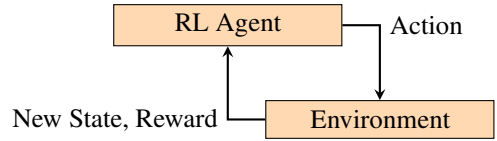

\subsection{Co-activation Graph Analysis}\label{sec:co}
The co-activation values are extracted over a set of inputs $\overline{S}$.
A co-activation value $W_{ij}^{kl}$ between a neuron $i$ in layer $k$ and a neuron $j$ in layer $l$ is defined as the correlation of
the activation values $A$ of the two neurons for a set of inputs $\overline{S}$~\cite[see Equation~\ref{eq:corr}]{DBLP:conf/kcap/SelaniT21}.
\begin{equation}\label{eq:corr}
    W_{ij}^{kl} = Corr (A(i, k, \overline{S}), A(j, k, \overline{S}))
\end{equation}
A co-activation graph is an undirected network where each node represents a neuron from the NN, and the connection weights indicate co-activation values~\citep{DBLP:journals/fgcs/HortaTLM21}.

\paragraph{PageRank.}
In graph theory, centrality measures quantify the importance of nodes within a graph.
The graph's domain and the chosen centrality metric determine the importance.
In the context of the co-activation graph, centrality measures can highlight neurons critical for the NN's performance~\citep{DBLP:journals/fgcs/HortaTLM21}.

The \emph{PageRank} centrality assigns an importance score to each node by considering both its direct connections and the importance of its neighbors~\citep{page1999pagerank}.
Let \( N \) be the total number of nodes in \( G \), and let \( \mathbf{PR} = [PR(n_1), PR(n_2), \dots, PR(n_N)]^\top \) be the PageRank vector initialized with \( PR(n_i) = 1 \) for all \( i \).
The PageRank is computed iteratively using~\citep{DBLP:conf/kcap/SelaniT21}:
\begin{equation}
PR(n_i) = \frac{1 - d}{N} + d \sum_{n_j \in \mathcal{N}(n_i)} \frac{A_{ji} \cdot PR(n_j)}{D(n_j)},
\label{eq:pagerank}
\end{equation}
where:
\begin{itemize}
    \item \( d \in (0,1) \) is the damping factor (typically \( d = 0.85 \)),
    \item \( \mathcal{N}(n_i) \) is the set of neighbors of node \( n_i \),
    \item \( A_{ji} = W_{ji} \) is the weight of the edge between nodes \( n_j \) and \( n_i \),
    \item \( D(n_j) \) is the weighted degree of node \( n_j \), defined as:
    \begin{equation}
    D(n_j) = \sum_{n_k \in \mathcal{N}(n_j)} A_{jk}.
    \label{eq:degree_centrality}
    \end{equation}
\end{itemize}
The iterative process continues until convergence, i.e., when \( \| \mathbf{PR}^{(t)} - \mathbf{PR}^{(t-1)} \| < \epsilon \) for a predefined \( \epsilon > 0 \). In the co-activation graph, a node with a high PageRank score corresponds to a neuron that is strongly correlated with many other influential neurons.

\paragraph{Louvain Community Detection.}

The Louvain community detection algorithm~\citep{DBLP:journals/corr/abs-2311-06047} is employed to identify community structures within the co-activation graph. This method optimizes the \emph{modularity} \( Q \) of the partitioning, which measures the density of links inside communities compared to links between communities. Modularity is defined as:
\begin{equation}
Q = \frac{1}{2m} \sum_{i,j} \left( A_{ij} - \frac{k_i k_j}{2m} \right) \delta(c_i, c_j),
\label{eq:modularity}
\end{equation}
where:
\begin{itemize}
    \item \( A_{ij} = W_{ij} \) is the weight of the edge between nodes \( i \) and \( j \),
    \item \( k_i = \sum_{n_j \in \mathcal{N}(n_i)} A_{ij} \) is the weighted degree of node \( i \),
    \item \( m = \frac{1}{2} \sum_{i,j} A_{ij} \) is the total weight of all edges in the graph,
    \item \( c_i \) is the community assignment of node \( i \),
    \item \( \delta(c_i, c_j) \) is the Kronecker delta, \( \delta(c_i, c_j) = 1 \) if \( c_i = c_j \) and \( 0 \) otherwise.
\end{itemize}
The modularity \( Q \) ranges between \( -1 \) and \( 1 \), where higher values indicate a stronger community structure. A high modularity implies that nodes are more densely connected within communities than between~them.

\section{Methodolodgy}
Our methodology consists of two main steps: generating a labeled dataset based on safety properties or other explainable RL or user-specified methods and applying co-activation graph analysis on the labeled dataset to interpret the NN policy.
The steps are detailed in the first two subsections, followed by a limitation analysis of our methodology.

\subsection{Labeled Dataset Generation}
In the first step, we create a dataset of states reflecting a user-specified safety property.
Given an MDP of the RL environment, a trained RL policy $\pi$, and a desired safety property,
we first incrementally build the induced DTMC of the policy $\pi$, and the MDP $M$ as~follows.

For every reachable state $s$ via the trained policy $\pi$, we query for an action $a = \pi(s)$. In the underlying MDP $M$, only states $s'$ reachable via that action $a \in A(s)$ are expanded. The resulting DTMC $D$ induced by $M$ and $\pi$ is fully deterministic, with no open action choices, and is passed to the model checker Storm for verification, yielding the \emph{exact} results concerning satisfying the safety property or violating it, and the states $\overline{S}$ belonging to the specific safety property~\citep{DBLP:conf/setta/GrossJJP22}.

Now, we have two options to proceed.

\paragraph{Option 1}
We label the entire dataset by associating each state $\overline{s} \in \overline{S}$ with the specific safety property.
In addition to this labeling, we also create alternative labeled datasets for comparative analysis. 
For instance, we may label the dataset according to a different safety property or specify a particular metric of interest (such as states with specific properties).
This enables us to explore policy behavior variations under different metrics.

\paragraph{Option 2}
By introducing other metrics, we can also classify each state $\overline{s} \in \overline{S}$ individually.
For example, we may categorize each state $\overline{s}$  as ``critical'' or ``non-critical'' based on the policy’s outputs.
These metrics can also be user-defined, allowing customization to reflect the policy's inner workings for the initial safety~property.

\subsection{Co-Activation graph Analysis}
We apply co-activation graph analysis on the labeled dataset to interpret the RL policy's neural network's internal structure and decision-making process.
For each label, we conduct the analysis separately, examining all data points associated with that label.
This allows us to understand how the network behaves differently across distinct labels.
By comparing the results across labels, we can identify label-specific influences within the NN and highlight structural and functional differences in its inner workings.

In this analysis, each neuron in the NN is represented as a node, and the connections between neurons are weighted by the statistical correlations in their activation patterns.
By examining this graph's structure and community relationships, we identify key neurons, rank influential state features, and assess the density of connections within and between neuron communities. 
We refer to the background section for details about the specific graph analysis algorithms (see Section~\ref{sec:co}).

These insights are crucial for interpretability, as they reveal decision-making pathways within the neural network, clarifying how specific inputs drive policy actions.

\paragraph{Co-activation Graphs in classification and sequential decision-making}
Co-activation graph analysis relies on a labeled dataset, with the labeling process differing between classification tasks and sequential decision-making.
In our RL safety setting, the dataset is generated via policy model checking and labeled by a user-defined function.

When combined with model checking, coactivation graph analysis provides more global insights than local explanation methods focusing on a single state.
At the same time, it offers a more fine-grained understanding than typical global explanation methods by revealing the policy's behavior in specific regions of the RL environment where safety properties are held.

\subsection{Advantages and Limitations}
RL policy co-activation graph analysis without safety properties is also possible for model-free RL environments (without rigorous model checking) by collecting states $\overline{s}$ for $\overline{S}$ by executing the policy in the environment and labeling them just via local RL explanation methods.

Our explainable RL safety method with co-activation graph analysis supports memoryless NN policies within modeled MDP environments, limited by its model checking for large state space and transition counts~\citep{DBLP:conf/setta/GrossJJP22}.
The co-activation graph analysis works with any layer in the NN architecture and supports labeled datasets of different sizes that can be found in labeled datasets for classification tasks~\citep{DBLP:journals/fgcs/HortaTLM21}.

While we built our work on top of the COOL-MC~\citep{DBLP:conf/setta/GrossJJP22}, we do not see any limitations in replacing this specific verification tool with other ones, such as \emph{MoGym}~\citep{DBLP:conf/cav/GrosHHKKW22}.

\section{Experiments}
In this section, we evaluate our proposed method and show that it is applicable in the context of explainable RL safety.
We begin by introducing the RL environments used in our experiments.
Next, we describe the trained RL policies.
We then explain the technical setup.
After that, we apply our method for co-activation graph analysis in various RL safety settings.
The first setting uses co-activation graph analysis for two different safety properties, the second setting uses it for an explainable RL method in the context of a specific safety~property.
The final subsection summarizes additional observations of applying co-activation graph analysis in RL safety.

\paragraph{Environments}
In the experiments, we use a taxi and a cleaning robot environment that are described below.

The \emph{taxi agent} has to pick up passengers and transport them to their destination without running out of fuel. The environment terminates as soon as the taxi agent does the predefined number of jobs or runs out of fuel. After the job is done, a new guest spawns randomly at one of the predefined locations~\citep{DBLP:conf/setta/GrossJJP22}.
For the first job, the passenger location and destination is always the same, afterwards the passenger location and destination is set to four locations randomly.
\begin{gather*}
S =
\{(x,y,passenger\_loc\_x,passenger\_loc\_y,\\ passenger\_dest\_x,\\ passenger\_dest\_y,fuel,\\done,on\_board,\\jobs\_done,done),\\...\} \\
Act = \{north,east,south,west,pick\_up,drop\}\\
Penalty = \begin{cases}
        0 \text{, if passenger successfully dropped.}
        \\
        21 \text {, if passenger got picked up.}
        \\
        21 + |x-passenger\_dest\_x| +\\ |y-passenger\_dest\_y|, \\ \text {if passenger on board.}
        \\
        21 + |x-passenger\_loc\_x| +\\ |y-passenger\_loc\_y| \text {, otherwise}
        \end{cases}
\end{gather*}

A \emph{robotic agent} cleans rooms while avoiding collisions and conserving energy.
Collisions refer to cleaning a room that has been cleaned by another entity.
The state includes room cleanliness, slipperiness, and the agent's battery level.
The agent can apply different cleaning options for different types of dirt that vary in their strength and effect on the environment.
The agent is rewarded for correct actions, and the environment terminates upon collisions, energy depletion, or cleaning an already clean room~\citep{gross2024enhancing}.

\begin{align*}
    S &= \{ \text{(dirt1, dirt2, energy, slippery\text{ }level},\\ & room\text{ }blocked), \dots \} \\
    Act &= \{ \text{next\text{ }room,charge\text{ }option1, charge\text{ }option2,} \\
    &\quad \text{clean1\text{ }option1, clean1\text{ }option2, clean2\text{ }option1,} \\
    &\quad \text{clean2\text{ }option2, all\text{ }purpose\text{ }clean, idle} \} \\
    \text{Reward} &= 
    \begin{cases}
        20 \cdot dirt*,\\ \text{if clean* operation for dirt* successful.} \\
        20 \cdot dirt1 \cdot dirt2,\\ \text{if all purpose clean operation successful.} \\
        20,\\ \text{if changing room correctly.}\\
        10,\\ \text{if idle when slippery level$>$0}\\ \text{an room not blocked.}\\
        10,\\ \text{if charging starts between}\\ \text{energy>0 and energy$\leq2$.}
        0,\\ \text{otherwise.}
    \end{cases}
\end{align*}

\paragraph{Trained RL policies}
For the taxi environment, we train all RL policies using the deep Q-learning algorithm~\citep{mnih2013playing}. In the \emph{taxi environment}, the trained RL policy received an average penalty of $-615$ over 100 episodes across 72,975 epochs.

For the cleaning robot environment, we train an RL policy using deep Q-learning~\citep{mnih2013playing} with 4 hidden layers of 512 neurons each. Training parameters were a batch size of 64, epsilon decay of 0.99999, minimum epsilon 0.1, initial epsilon 1, $\gamma$ 0.99, and target network updates every 1024 steps. The policy achieved an average reward of 67.8 over 100 episodes in 27,709 epochs.

\paragraph{Technical setup}
We executed our benchmarks in a docker container with 16 GB RAM, and an AMD Ryzen 7 7735hs with Radeon graphics × 16 processor with the operating system Ubuntu 20.04.5 LTS.
For model checking, we use Storm 1.7.1 (dev).
The code can be found on \url{https://github.com/LAVA-LAB/COOL-MC/tree/co_activation}.

\subsection{Comparative Analysis of Trained RL Policies Across Different Safety Scenarios}\label{sec:safety}
In this experiment, we show, using the \emph{taxi} environment, that it is possible to use co-activation graph analysis with model checking to gain safety insights into the trained RL policy and observe NN inner workings for different safety~properties.

\paragraph{Setup} In this experiment, we compare the co-activation graphs of a policy that differs for the datasets associated with the safety property of finishing with a reachability probability of 1 one job ($P_{=1}(\lozenge jobs=1)$) vs. finishing two jobs ($P_{=1}(\lozenge jobs=2)$).
This can give us, for instance, insights into what features are more relevant in the beginning of the taxi policy execution compared to later steps in the environment.

\paragraph{Execution}
We first create the two labeled datasets by building the formal model for each and verify each model with the corresponding PCTL queries.
The labeled dataset for safety property $P_{=1}(\lozenge jobs=1)$ has 12 data points and the labeled dataset for safety property $P_{=1}(\lozenge jobs=2)$ has 206 data points.

We apply the PageRank algorithm to rank all neurons in the neural network and use the Louvain community detection algorithm to identify neuron communities and calculate modularity values.

\paragraph{Results}
\begin{figure}
    \centering
    \includegraphics[width=1.1\linewidth]{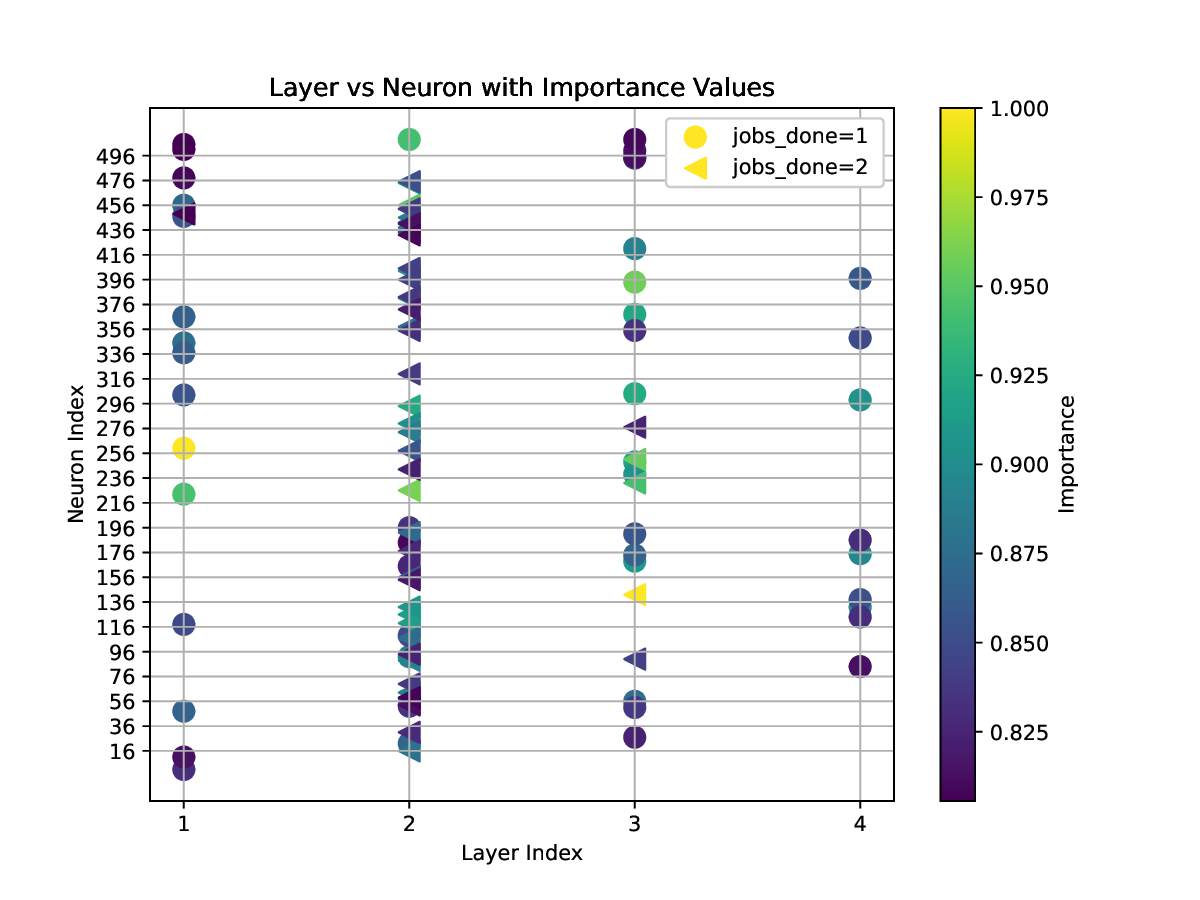}
    \caption{The 50 most significant neurons identified for each safety property.}
    \label{fig:centrality_safety}
\end{figure}
In Figure~\ref{fig:centrality_safety}, we observe that the 50 most important neurons differ across safety properties.
The most important neuron for the safety property of finishing one job is in hidden layer 1, while it is in hidden layer 3 for finishing two jobs.

The most important state features for the property of finishing two jobs are the passenger’s destination and the number of jobs completed, and indeed, when you prune the outgoing connections from these feature neurons~\citep{SafetyPruningESANN2024}, the safety property gets violated, and the reachability probability of finishing two jobs is only $0.25$ indicating that it just randomly selects one out of the four different passenger targets.

For finishing one job, the most significant features are the fuel level, the passenger’s x-coordinate destination, and the number of jobs completed.
And indeed, when pruning these feature neurons, the safety property gets violated and the reachability probability of finishing one jobs drops to zero.
The initial location of the passenger is irrelevant for the first job because the starting location and destination are consistent in the environment; however, it becomes relevant for subsequent passengers, which may also explain the importance of the number of jobs completed.

Community analysis reveals stronger interactions among neurons, with a modularity of $0.29$ for $P_{=1}(\lozenge jobs=1)$ and $0.25$ for $P_{=1}(\lozenge jobs=2)$. The community overlap is approximately $97\%$, indicating that the same regions of the neural network process both properties.

\subsection{Critical vs. Non-Critical State Analysis for a Specific Safety Scenario}
In this experiment, we target a specific safety property for applying co-activation graph analysis.
We categorize the states using a local explainable RL method to label the dataset.
This approach demonstrates that integrating model checking and established local explainable RL methods with co-activation graph analysis can yield deeper insights into neural network policy decision-making. Experiments are again performed on the \emph{taxi} environment.

\paragraph{Setup}
We focus on a dataset of states linked to the safety property of completing two jobs with a reachability probability of 1 ($P_{=1}(\lozenge jobs=2)$).
Labeling is performed through critical state classification, where a threshold of 100 is set for the distance between the highest and lowest predicted Q-values of the policy.
Each state is labeled as critical if this Q-value distance meets or exceeds the~threshold.

\paragraph{Execution}
We collect the states associated with $P_{=1}(\lozenge jobs=2)$ and label each state as either critical or non-critical based on its classification criteria.
In total, we got 206 data points, where 22 data points are labeled as critical and 184 as non-critical.

We apply the PageRank algorithm to rank all neurons in the neural network and use the Louvain community detection algorithm to identify neuron communities and calculate modularity values.

\begin{figure}
    \centering
    \includegraphics[width=1.1\linewidth]{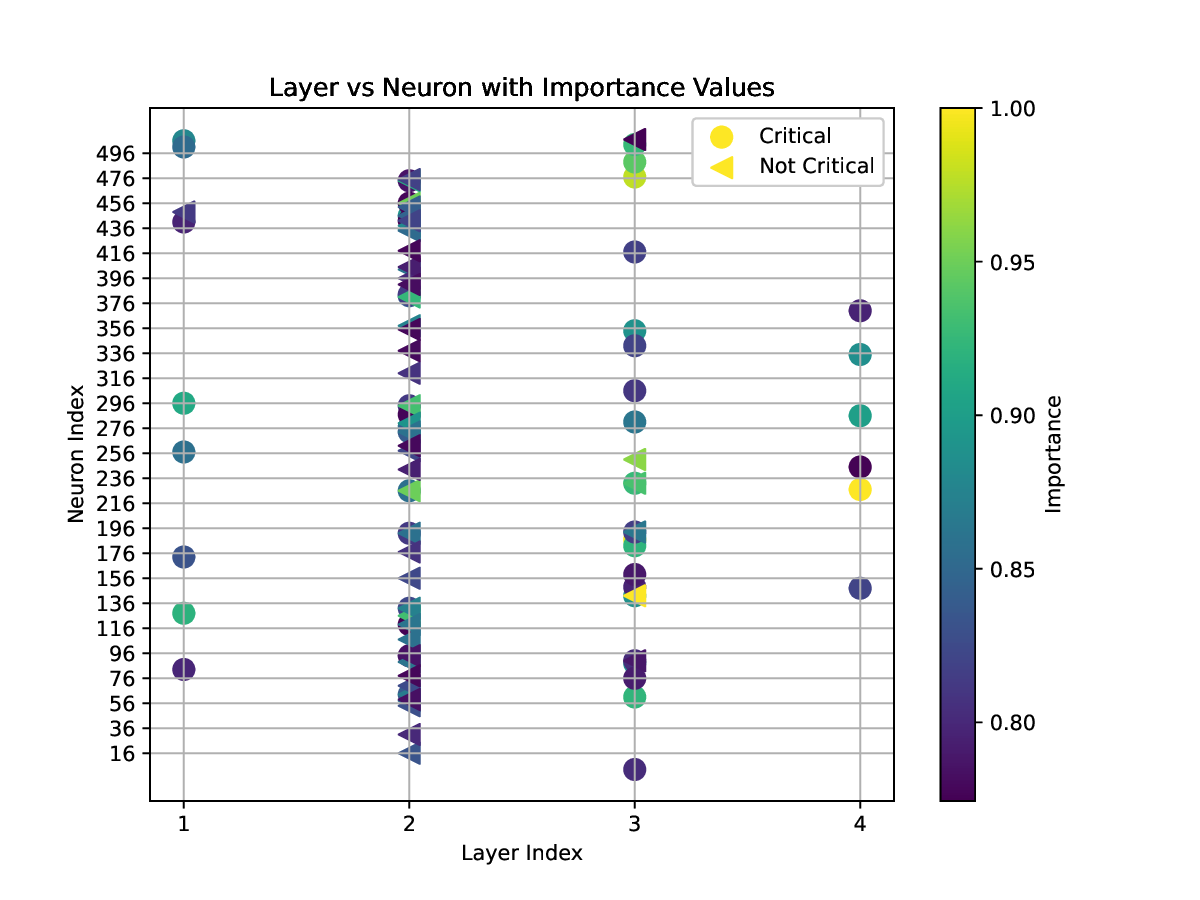}
    \caption{The 50 most significant neurons for the safety property $P_{=1}(\lozenge jobs=2)$ identified for critical and non-critical states.}
    \label{fig:centrality_critical}
\end{figure}

\paragraph{Results}
In Figure~\ref{fig:centrality_critical}, we observe the 50 most important neurons for critical and non-critical states.
We observe that the most influential neuron is located in layer 3 for non-critical states.
In contrast, for critical states, it is in layer 4, providing insights into the neural network's internal processing.

In analyzing the key features from the centrality analysis, we find that the top three features for critical states involve the passenger's destination and whether the passenger is on board.
In contrast, the most significant features for non-critical states are the passenger's destination and current x-coordinate.

Community analysis reveals stronger interactions among neurons for critical states, with a modularity of 0.25 compared to 0.23 for non-critical states. Community overlap is approximately 93\%, indicating that both critical and non-critical states are processed in largely the same neural network regions.

\subsection{Additional Observations}
Next to the two different ways to apply co-activation graph analysis in the context of RL safety, we made the following observations.

\paragraph{Common observations in both environments}
Interestingly, hidden layer 2 appears to be the most relevant in both experiments on average. This suggests that most of the decision-making process is completed at this stage, with the final layer primarily serving to route the decision to the appropriate action.

\paragraph{Comparative method analysis}
Further, we compared our approach with an alternative explainable RL safety approach from~\cite{SafetyPruningESANN2024}.

In our experiment in Section~\ref{sec:safety}, we show that the most relevant features for satisfying the safety property $P_{=1}(\lozenge jobs=2)$ are the passenger’s destination and the number of jobs completed, and indeed, when you prune the outgoing connections from these feature neurons, the safety property gets violated, and the reachability probability of finishing two jobs is only $0.25$ indicating that it just randomly selects one out of the four different passenger targets.

For finishing one job, the most significant features are the fuel level, the passenger’s x-coordinate destination, and the number of jobs completed.
Indeed, when pruning these feature neurons, the safety property gets violated, and the reachability probability of finishing one job drops to zero.

These findings conform that our method correctly identifies significant and highly relevant neurons for the safety properties.

\paragraph{Different RL environments} Finally, we applied co-activation analysis to the cleaning robot environment. 
Here, for a labeled dataset for the safety property $P_{=0.6} (\lozenge energy=0)$ and $P_{=0.02}(\lozenge \text{wrong room switch})$, we observe that the communities overlap in $0.95$ of the cases while the feature importance ranking is the same.
The modularity for the first safety property $P_{=0.6} (\lozenge energy=0)$ is $0.37$ an for the second property $P_{=0.02}(\lozenge \text{wrong room switch})$ is $0.35$.

While we focus in our investigation mainly on the taxi environment, this experiment confirms that our method is similarly applicable to other environments.

\section{Conclusion}
In this paper, we introduced a methodology that integrates RL policy model checking~\citep{DBLP:conf/setta/GrossJJP22} with co-activation graph analysis to improve the explainable safety of RL policies.
By generating labeled datasets through model checking and local explainable RL methods, we extended co-activation graph analysis~\citep{DBLP:conf/dexaw/HortaM19} to apply it within RL safety.
Our approach enables examining NN policies by analyzing neuron activation patterns in states associated with specific safety properties and local explainable RL method results.

For future work, we plan to examine how co-activation graph analysis can be applied within multi-agent RL settings~\citep{DBLP:journals/aamas/ZhuDW24} or to be used for safe NN policy pruning~\citep{SafetyPruningESANN2024}.

\bibliographystyle{apalike}
{\small
\bibliography{example}}

\end{document}